%% file: main.tex
\pdfoutput=1
\documentclass[11pt]{article}

\usepackage[preprint]{acl}
\usepackage{graphicx}
\usepackage{pgffor}
\usepackage{graphicx}
\usepackage{multirow}
\usepackage{marvosym}
\usepackage{footmisc}
\usepackage{booktabs} % 用于更美观的表格线
\usepackage{times}
\usepackage{amsthm}
\usepackage{algorithm}
\usepackage{algpseudocode}
\usepackage{latexsym}
\usepackage{float} % 引入float宏包，用于禁止浮动体浮动
\usepackage{amsmath}
\usepackage[T1]{fontenc} % For proper rendering and hyphenation of words containing Latin characters (including in bib files)
\usepackage[utf8]{inputenc} % This assumes your files are encoded as UTF8
\usepackage{microtype}
\usepackage{inconsolata}
\usepackage{amsfonts}
\usepackage{graphicx}
\usepackage{hyperref}       % hyperlinks
\usepackage{url}            % simple URL typesetting
\usepackage{nicefrac}       % compact symbols for 1/2, etc.
\usepackage{xcolor}         % colors
\usepackage{tabularx}
\usepackage{adjustbox}
\usepackage{tabularray}
\usepackage{wrapfig}
\usepackage{longtable}
\usepackage{tabularx}
\usepackage{listings}
\usepackage{verbatim}
\usepackage{url}
\usepackage{ulem}
\usepackage{xcolor}
\definecolor{darkgreen}{RGB}{0, 0.392, 0} 
\title{DSVD: Dynamic Self-Verify Decoding for Faithful Generation in Large Language Models}
\author{
 \textbf{YiQiu Guo\textsuperscript{1,2}},
 \textbf{Yuchen Yang\textsuperscript{2,4}},
 \textbf{Zhe Chen\textsuperscript{2,3}},
 \textbf{Pingjie Wang\textsuperscript{2,3}},
\\
 \textbf{Yusheng Liao\textsuperscript{2,3}},
 \textbf{Ya Zhang\textsuperscript{2,3}}$\textsuperscript{\Letter},
 \textbf{Yanfeng Wang\textsuperscript{2,3}},
 \textbf{Yu Wang \textsuperscript{2,3}}$\textsuperscript{\Letter},
\\
 \textsuperscript{1}Fudan University,
 \textsuperscript{2}Shanghai AI Laboratory,
 \textsuperscript{3}Shanghai JiaoTong University,
\\
 \textsuperscript{4}University of Science and Technology of China,
\\
 \small{
   \Letter: Corresponding author.
 }
}

\begin{document}
\maketitle

\input{sections/intro}

\input{sections/method}

\input{sections/evaluation}

\bibliography{custom}

\newpage
\input{sections/appendix}

\end{document}

%% file: sections/intro.tex
\begin{abstract}
% \footnotetext{\Letter: Corresponding author.}
The reliability of large language models remains a critical challenge, particularly due to their susceptibility to hallucinations and factual inaccuracies during text generation.
Existing solutions either underutilize models' self-correction with preemptive strategies or use costly post-hoc verification. To further explore the potential of real-time self-verification and correction, we present Dynamic Self-Verify Decoding (DSVD), a novel decoding framework that enhances generation reliability through real-time hallucination detection and efficient error correction. DSVD integrates two key components: (1) parallel self-verification architecture for continuous quality assessment, (2) dynamic rollback mechanism for targeted error recovery.
Extensive experiments across five benchmarks demonstrate DSVD's effectiveness, achieving significant improvement in truthfulness (Quesetion-Answering) and factual accuracy (FActScore). Results show the DSVD can be further incorporated with existing faithful decoding methods to achieve stronger performance.
Our work establishes that real-time self-verification during generation offers a viable path toward more trustworthy language models without sacrificing practical deployability. 
\end{abstract}

\section{Introduction}

\begin{figure}[t!]
    \centering
    \includegraphics[width=1\linewidth]{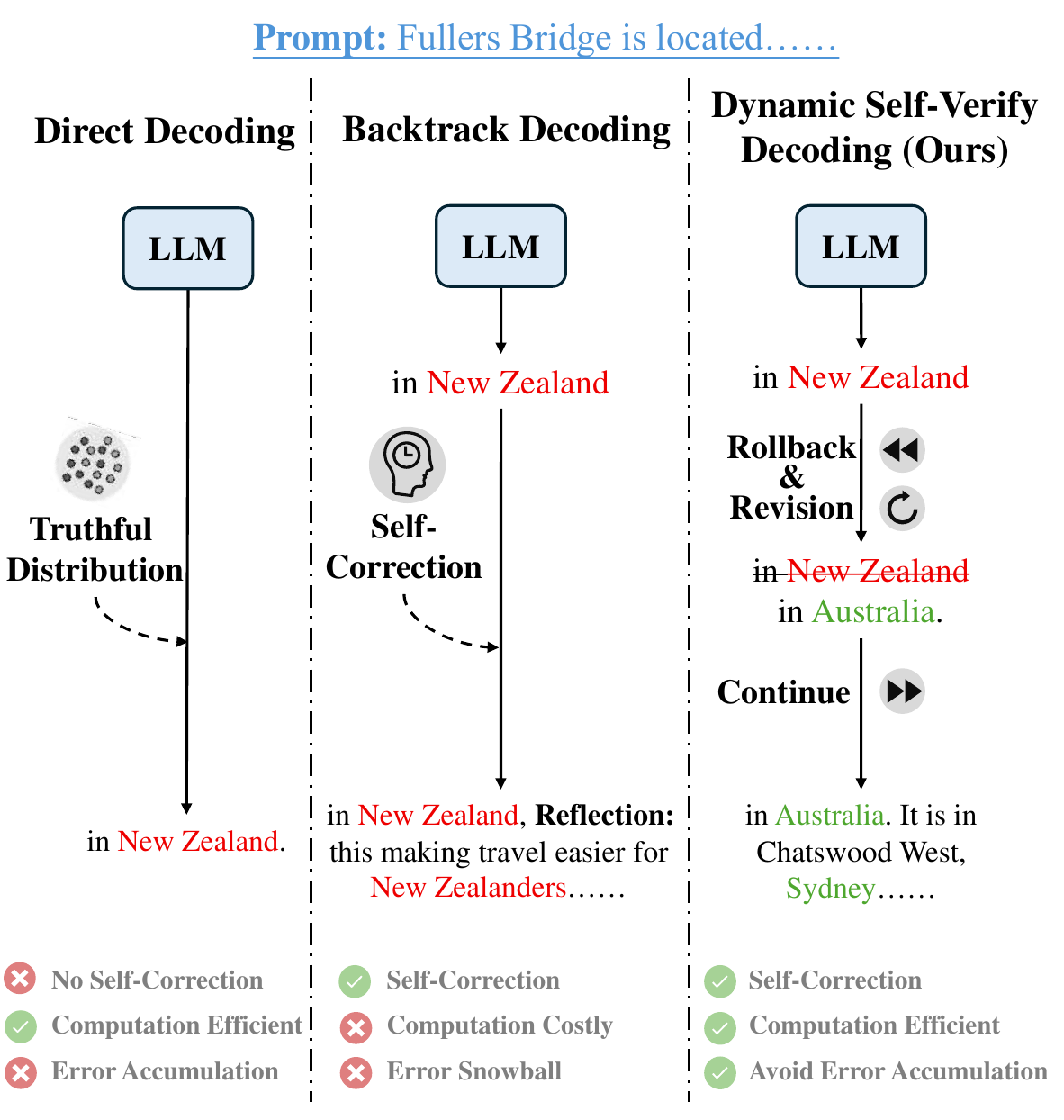}
    \caption{Comparative analysis of different decoding strategies: (a) Direct decoding leaves existing errors unexploited. (b) Baseline backtrack decoding propagates geographic hallucination and incurs high computation costs. (c) Our dynamic approach corrects \textit{"New Zealand"→"Australia"} with minimal overhead.}
    \label{fig:overview}
\end{figure}
Large Language Models (LLMs) have demonstrated remarkable capabilities across various natural language processing (NLP) tasks, including question-answering, summarization, and conversation generation~\cite{openaiGPT4TechnicalReport2024, deepseek-aiDeepSeekR1IncentivizingReasoning2025, touvronLLaMAOpenEfficient2023}. Despite their impressive performance, these models frequently suffer from reliability issues manifested through hallucinations and factual inaccuracies~\cite{LanguageModels-2022a, xiongCanLLMsExpress2023, gekhmanDoesFineTuningLLMs2024, liSurveyHonestyLarge2024}. These deficiencies pose significant practical concerns as users may unwittingly trust erroneous information presented in the models' confident and coherent outputs.

Recent advancements in faithful generation have shifted focus towards inference-stage interventions~\cite{liangInternalConsistencySelfFeedback2024, luoSEDSelfEvaluationDecoding2024, chenInContextSharpnessAlerts2024}. Researchers recognize that even models containing factual knowledge during pre-training often fail to access this information during generation reliably. Decoding-time adjustment strategies present a promising direction, offering more cost-effective solutions compared to supervised fine-tuning (SFT), which requires substantial computation, or retrieval-augmented generation (RAG), which necessitates an external knowledge base. As illustrated in Figure~\ref{fig:overview}, existing faithful generation approaches can be categorized into two paradigms: 
\textbf{Direct Decoding} methods (e.g., ITI~\cite{liInferenceTimeInterventionEliciting2023}, DoLa~\cite{chuangDoLaDecodingContrasting2024}, TruthX\cite{zhangTruthXAlleviatingHallucinations2024}) steer model outputs toward truthful directions by manipulating internal representations, leveraging the model's inherent truthful priors. While effective, these approaches fail to leverage the model's ability for self-correction of errors and reflective reasoning, leaving the model powerless against error accumulation.
\textbf{Backtracking Decoding} methods (e.g., Self-Refine~\cite{madaanSelfRefineIterativeRefinement2023}, Reflexion~\cite{shinnReflexionLanguageAgents}) employ post hoc verification of generated content, but existing implementations suffer from significant computational overhead and vulnerability to error accumulation, where initial errors propagate into subsequent generations through self-reinforcing mechanisms.
\begin{table}
    \centering
    \begin{tabular}{lcc}
    \toprule
         \begin{tabular}[c]{@{}c@{}}Dataset\end{tabular}&  \begin{tabular}[c]{@{}c@{}}Probing\\w/o Response\end{tabular}&  \begin{tabular}[c]{@{}c@{}}Probing\\w/ Response\end{tabular}\\
         \midrule
         SciQ& 64.86&\textbf{87.21}\\
         CoQA&62.98& \textbf{76.88}\\
         TriviaQA& 68.33& \textbf{75.66}\\
         % MedMCQA& 65.16& \textbf{77.08}\\
         % MedQA& 65.80& \textbf{80.82}\\
         \bottomrule
    \end{tabular}
    \caption{Experiment results of validating our insights: delayed awareness of hallucinations, the metric is AUROC. More details and analysis are in Appendix~\ref{app:insights}}.
    \label{tab:insights}
\end{table}

To address these limitations, we propose \textbf{D}ynamic \textbf{S}elf-\textbf{V}erify \textbf{D}ecoding (DSVD), a novel decoding strategy that incorporates real-time self-verification with dynamic rollback mechanisms. Our approach builds on key insights: (1) \textbf{Delayed Awareness of Hallucinations}: models demonstrate superior ability in detecting existing errors compared to preemptively preventing them, as shown in~\ref{tab:insights}, and (2) \textbf{Local Error Correction is more Efficient}: localized rollback enables error correction at their source, offering higher efficiency than global rewriting. The method mirrors human behavioral patterns: speculating, verifying, and refining the consequences before reaching a conclusion. More specifically, the framework operates through two components: 1) Fine-grained hallucination detector trained on model-generated pseudo-labels; and 2) Parallel self-verification and dynamic rollback mechanism enabling real-time hallucination detection and error correction.

Our experimental evaluation across multiple LLM architectures (LlaMA-2, LlaMA-3, Qwen-2.5) and benchmarks (TruthfulQA, StrQA, SciQ, EntityQuestions, FActScore) demonstrates consistent improvements in truthfulness and factual accuracy while maintaining computational efficiency. Notably, DSVD shows complementary benefits when combined with existing direct generation methods, suggesting orthogonal mechanisms of action. Our key contributions include:
\begin{itemize}
    \item We propose a straightforward yet intuitive semi-supervised hallucination labeling approach for fine-grained self-feedback.
    \item We propose Dynamic Self-Verify Decoding: a novel decoding strategy that enables parallel self-verification and dynamic self-correction.
    \item   Comprehensive experiments across diverse LLMs and evaluation metrics reveal consistent performance improvements of DSVD.
\end{itemize}

\section{Related Work}

\subsection{Faithful Decoding}
In recent years, a series of studies have focused on leveraging truthful distribution to intervene in the model's next-token prediction. Some research has explored directing the model's generation towards a "more truthful" direction through representation editing. ITI~\cite{liInferenceTimeInterventionEliciting2023} trains probing heads to identify a set of more truthful attention heads and enhances the weights of these heads during inference. TrFr~\cite{chenTruthForestMultiScale2024a} proposed the application of multi-dimensional orthogonal probes, which effectively extract features from both truthful and non-truthful texts to better identify effective attention heads. TruthX~\cite{zhangTruthXAlleviatingHallucinations2024} not only targets attention heads but also latent states in the forward feedback layer. By separately mapping these states using truthful and semantic encoders.

Another line of research investigates contrastive decoding for faithful generation. The pioneering work by~\cite{liContrastiveDecodingOpenended2023} introduced Contrastive Decoding, which selects optimal tokens by contrasting probability distributions from expert and amateur models. Building on this foundation, DoLa~\cite{chuangDoLaDecodingContrasting2024} enhanced the framework by incorporating intermediate layer representations, thereby improving early-stage reasoning consistency and pre-answer alignment through its Decoding-by-Contrasting-Layers mechanism. SLED proposed by~\cite{zhangSLEDSelfLogits2024} integrates latent knowledge into logits via single-step gradient-like operation instead of replacing original outputs in DoLa during inference. 

\paragraph{Our Innovation:} The direct decoding methods mostly intervene before the model predicts the next token, thus the model's self-awareness and self-feedback capabilities regarding hallucinations are unexploited, while DSVD intervene after the model encounter hallucination and thus fully utilize the self-reflection ability of large language models.  

\subsection{Self Feedback}
\begin{figure*}[t!]
    \centering
    \includegraphics[width=1\linewidth]{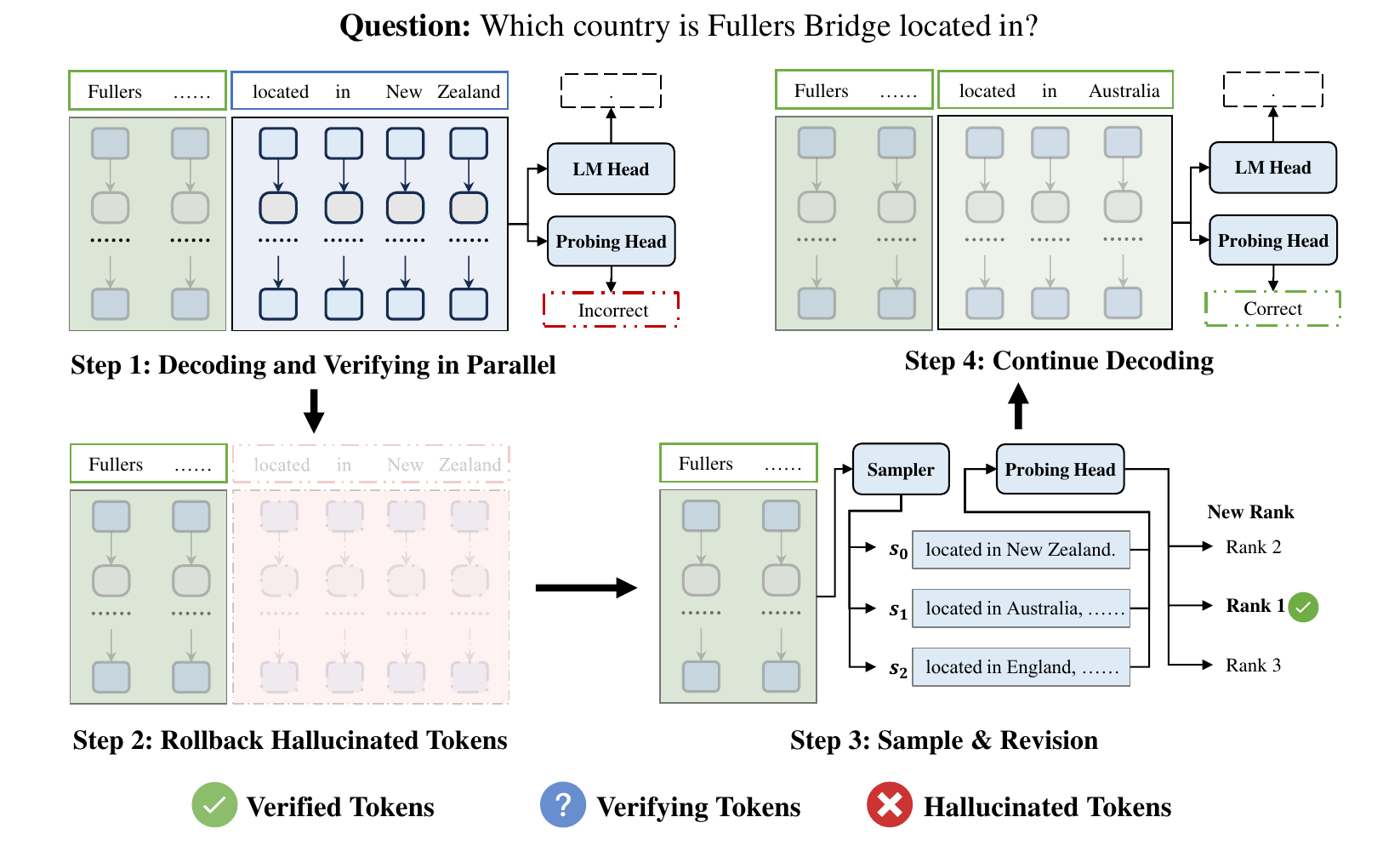}
    \caption{Illustration of the Dynamic Self-Verify Decoding Framework: \textbf{Step 1}: Parallel hallucination detection through trained probing heads, operating concurrently with the LM Head's next-token prediction; \textbf{Step 2}: Dynamic rollback to pre-hallucination positions upon error detection; \textbf{Step 3}: Sample candidate continuation with probing-head-derived penalty terms for re-ranking; \textbf{Step 4}: Resumption of the decoding process with revised token sequences.}
    \label{fig:pipeline}
\end{figure*}
Studies on self-feedback utilize the Large Language Model itself as a critic, enabling the model to generate feedback on its responses and further refine those responses based on the generated feedback. Self-Refine~\cite{madaanSelfRefineIterativeRefinement2023} simply uses the LLM in SelfEvaluate(·) to generate textual feedback. Reflexion~\cite{shinnReflexionLanguageAgents} makes progress by regarding iterative refinement as Verbal Reinforcement Learning without weight updates. Self-Correct~\cite{welleckGeneratingSequencesLearning2022} uses the same framework but trains a Corrector model for better feedback. Yet, due to not being task-agnostic and the need for training, it reduces the framework's flexibility.

\paragraph{Our Innovation:} Traditional self-feedback approaches incur significant overhead by operating through textual critique generation. DSVD circumvents these limitations through two innovations: (1) direct utilization of internal consistency signals as implicit feedback, avoiding costly text generation cycles; (2) localized correction via hidden state rollback instead of full-sequence regeneration reducing computation cost compared to prior methods. More comparisons and discussions with other related work can be found in Appendix~\ref{app:more_rw}

%% file: sections/method.tex
\section{Dynamic Self-Verify Decoding}

The dynamic self-verify decoding pipeline has two main steps. First, create a specialized hallucination detector. This detector analyzes the LLM's internal states to measure its prediction confidence. Second, use the hallucination detector during decoding. It serves as an alert for when the model might hallucinate and as a penalty term when the model samples to improve predictions. This section first formalizes the detector's construction process and then explains in detail how we use it as an indicator and penalty term during the model's decoding process.

\subsection{Train Fine-grained Hallucination Detector}

Inspired by the recent work on the internal consistency of large language models\cite{liangInternalConsistencySelfFeedback2024}, we create a specialized fine-grained hallucination detector for each large language model in a semi-supervised manner. We train a group of probing heads with LLM's internal states using a certain number of self-generated samples. The hallucination detector is created in the following steps:

\paragraph{Fine-Grained Train Data Construction} First, we select the training split of a general domain question-answer bank EntityQuestions~\cite{sciavolinoSimpleEntityCentricQuestions2021} with correct standard answers. Initially, the model is utilized to generate responses. Subsequently, the Rouge-L metric~\cite{linROUGEPackageAutomatic2004} is computed between the generated response and the ground truth. To avoid the influence of noise in the data, we differentiate between correct and incorrect responses by identifying those with an F1-measure value of Rouge-L greater than 0.8 and less than 0.2 respectively. For correct responses, we simply assign a label of zero to each token within them. For incorrect responses, we identify hallucinated points by calculating each token's conditional probability of generating ground truth tokens. Specifically, if a token position shows a significantly higher probability of producing ground truth tokens compared to other positions but fails to do so, we mark it as a hallucination point. We will elaborate on this process in detail below.

Consider a model's incorrect response $X=(x_0,x_1,x_2,\cdots,x_N)$, where $N$ indicates the number of tokens within the response and $x_i$ is the individual token it contains. Similarly, the ground truth tokens are identified as ~$G=(g_0,g_1,g_2,\cdots,g_M)$ with $M$ tokens and $g_i$ represent tokens in it. For each response, we calculate the score of hallucination occurrence at the position $i$ as:
\begin{equation}
\mathcal{P}_i^{gt}=\sum_{j = 0}^{M}\log(p(g_j|x_0:x_{i-1},g_0\cdots g_{j-1}))
\label{eq:data1}
\end{equation}
$\mathcal{P}_i^{gt}$ is the score of hallucination occurrence, $i$ represents the position index of the token. $p(g_j|x_0 \cdots x_{i-1},g_0\cdots g_{j-1})$ is the conditional probability of the $j$-th ground truth tokens with $i$ response tokens as its prefix. Then we assign token-level labels $y_i$ for each token within the response in: 
\begin{equation}
y_i=\begin{cases}
0, & \text{if } i<argmax(\mathcal{P}_{gt})\\
1, & \text{if } i=argmax(\mathcal{P}_{gt})\\
-1, & \text{if }i>argmax(\mathcal{P}_{gt})
\end{cases}
\label{eq:data2}
\end{equation}
The highest $\mathcal{P}_i^{gt}$ in the response is selected as the hallucination occurrence point of the response. We construct the training dataset by splitting it into correct and hallucinated responses in a 50/50 ratio.

\paragraph{Model Architecture and Training Detail} After extensive experiments, we use a combination of $L$ probing heads to predict fine-grained hallucinations~\footnote{We experimented with various probing-head architectures and present the detailed results in the Appendix~\ref{app:probing_head}.}. Each of these probing heads is a two-layer MLP with a binary classification output, denoted as $\phi=(\phi^0, \phi^1,\cdots, \phi^L)$. During the forward process of LLM, we save the hidden states output by all model layers, represented as $H = (h^0,h^1,\cdots,h^L)$. We calculate probing logits for each layer, average them across all layers and apply a softmax function to obtain the binary probability $z_i$, expressed as:
\begin{align}
z_i&=softmax(\frac{1}{L}\sum^{L}_{l=0}\phi^l(h^l_i))
\label{eq: probing}
\end{align}
where $z_i=(z_i^{hallu},z_i^{correct})$ are the binary probing probability of each token at position $i$. We utilize the focal loss during training, which has a form of:
\begin{equation}
FL(z_i^t)=-(1-z_i^t)^{\gamma}\log(z_i^t)
\label{eq:focal_loss}
\end{equation}
where $t$ is the class index, $z_i^t$ is the probing probability for a token at position $i$ and $\gamma$ is the focusing parameter. During the training, we use the AdamW optimizer with a learning rate of 1e-4, we set $\gamma=2$ in Eq.\ref{eq:focal_loss} and train each model for 10 epochs.
\begin{algorithm}[t!] 
    \caption{Dynamic Self-Verify Decoding}  
    \label{alg:DSVD}  
    \begin{algorithmic}[1]  
        \State \textbf{Input: }LLM $\theta$, Probing heads $\phi$, Inputs $x$, rollback size $r$, sample length $m$, search width $k$, penalty intensity $\alpha$
        \State \textbf{Initialize: } generated sequence $s \leftarrow x$, current position $t \leftarrow |x|$, sliding window $\mathcal{W} \leftarrow \emptyset$
        
        \While{$t < t_{\text{max}}$ \textbf{and} $x_t \neq \text{<EOS>}$}
            \State Compute LM Probabilities $p_{t+1} = \theta(h_t)$
            \State Compute $z_t = \phi(h_t)$ via Eq.~\ref{eq: probing}
            \State $\mathcal{W} \leftarrow \mathcal{W} \cup \{z_t^{hallu}\}$ \label{line:window_update}
            
            \If{$\exists z_i \in \mathcal{W}: z_i^{hallu} > 0.5$} \label{line:check_condition}
                \State Rollback to position $x_{t-r}$ \label{line:rollback}
                \State Generate candidates $\mathcal{S}$ 
                \For{each candidate $s_j \in \mathcal{S}$} \label{line:scoring_loop}
                    \State Compute $f(s_j)$ via Eq.~\ref{eq:scoring} \label{line:compute_score}
                \EndFor
                
                \State Update $s \leftarrow s_{\text{best}}$ from Eq.~\ref{eq:best_selection} \label{line:update_seq}
                \State $t \leftarrow |s|$, $\mathcal{W} \leftarrow \emptyset$ \label{line:reset_counter}
            \Else
                \State Append $x_{t+1} = \arg\max p_{t+1}$ to $s$ \label{line:normal_decode}
                \State $t \leftarrow t + 1$ \label{line:increment}
            \EndIf
        \EndWhile
       \State \textbf{Return: } Generated sequence $s$
    \end{algorithmic}  
\end{algorithm}

\subsection{Decoding with Dynamic Self-Verification}

\paragraph{Decoding and Verifying in Parallel} During inference, our framework enables real-time hallucination detection by leveraging the trained probing heads and the LLM's intermediate hidden states. As illustrated in Figure~\ref{fig:pipeline}, the probing heads share the LLM's internal states with the language modeling head, enabling parallel computation of: 1) next token prediction via the LM head; 2) probing probability via Eq.~\ref{eq: probing}. This architectural design introduces negligible latency (measured at only 5\% extra latency in our experiments) as both components utilize the same hidden states.
\paragraph{Dynamic Rollback Mechanism} We implement the dynamic rollback mechanism by setting a sliding window that moves along with the currently predicted token with a configurable window size $r$:
\begin{equation}
\mathcal{W} = \{z_{t-r+1}, ..., z_t\}
\label{eq:window}
\end{equation}
where $t$ stands for the current generation length and $z_i$ is the probing probability in Eq. \ref{eq: probing}. The system triggers rollback when any element in $\mathcal{W}_t$ exceeds the threshold:
\begin{equation}
\exists z_i \in \mathcal{W} : z_i^{hallu} > 0.5 \Rightarrow \text{Rollback to }x_{t-r}
\label{eq:rollback_condition}
\end{equation}
The dual motivation for this design stems from our key observations:
\begin{enumerate}
    \item \textbf{Semantic Completeness Requirement}: Individual tokens lack sufficient semantic context for reliable hallucination detection. For instance, consider the partial generation "\textit{locate in New ZeaLand}" – the substring "\textit{locate in New}" may appear anomalous but requires subsequent tokens for proper validation.
    
    \item \textbf{Delayed Error Identification}: Through controlled experiments (see Section~\ref{sec:sensitivity_analysis}), we discovered that LLMs typically recognize their own errors a few tokens after the initial mistake. The sliding window mechanism accommodates this inherent latency while maintaining computational efficiency.
\end{enumerate}

\paragraph{Probing probability as A Penalty} Following rollback operations, we employ a sampling algorithm (we use beam search by default) to generate $k$ candidate continuations $\mathcal{S} = \{s_1, s_2, ..., s_k\}$ of length $m$ for correction. The probing probabilities $z^{hallu}$ are incorporated as penalty terms in the scoring function to prioritize candidates with lower hallucination risk. For each candidate sequence containing tokens $s_i$=$(x_{t_0}, ..., x_{t_0+m})$, where $t_0$ denotes the rollback position, we compute the penalized log-probability score:
\begin{equation}
f(s_j) = \sum_{i=t_0}^{m} \left[ \log(p(x_i | x_{<i})-\alpha \log(z^{hallu}_i )\right]
\label{eq:scoring}
\end{equation}
where $p(x_i |x_0 \cdots x_{i-1})$ represents the standard language modeling probability, and $\alpha \in \mathbb{R}^+$ controls the penalty intensity inspired by contrastive decoding approaches \cite{obrienContrastiveDecodingImproves2023}. The optimal continuation $s_{\text{best}}$ is selected through:
\begin{equation}
s_{\text{best}} = \underset{s_j \in \mathcal{S}}{\arg\max}\, f(s_j)
\label{eq:best_selection}
\end{equation}

%% file: sections/evaluation.tex
\begin{table*}[t!]
    \centering
    \begin{tabular}{l|ccccccc}
    \toprule
 & \multicolumn{3}{c}{\textbf{TruthfulQA}} &\multicolumn{3}{c}{\textbf{Question Answering}} &\textsc{\textbf{FActScore}}\\
        h\
        Model & Truth (\%) & Info (\%) & T*I (\%)  &StrQA&SciQ &EntQ&Score 
\\
        \midrule
        llama2-7b-chat&36.9 &86.2 &31.9 & 63.6& 59.8& 29.3&32.6 
\\
        \ + ITI       &41.7 &77.2 &32.4 & 55.7& 41.7& 19.8&22.6 
\\
        \ + DoLa&42.1 & \textbf{98.3}& 41.4& 62.1& 61.3& 29.5&32.7 
\\
 \ + TruthX& \textbf{61.1}& 74.1& 45.2& 57.6& 55.0& 25.7&32.1 
\\
        \ + Self-Refine& 39.4& 93.6& 36.9& 66.2& 61.2& 29.7&32.9 
\\
        \ + \textbf{DSVD(Ours)} & 56.3& 85.9& \textbf{48.4}& \textbf{67.7}& \textbf{61.8}	& \textbf{30.7}&\textbf{33.3} 
\\
        \midrule
        llama3-8b-it& 61.8& 80.4& 49.7& 77.2& 65.1& 36.6&35.9 
\\
        \ + ITI& \textbf{65.5}& 78.4& 51.3& 71.2& 63.2& 36.0&31.1 
\\
        \ + DoLa& 62.2& 82.0& 51.0& 76.9& 65.4& 36.6&36.4 
\\
        \ + Self-Refine& 62.7& 82.1& 51.5& 69.3& 65.4& 36.8&36.9 
\\
        \ + \textbf{DSVD(Ours)} & 64.5& \textbf{81.0}& \textbf{52.3}&\textbf{ 77.7}& \textbf{66.4}& \textbf{37.1}&\textbf{37.7} 
\\
        \midrule
 qwen2.5-7b-it& 86.3& 32.9& 28.4& 77.6& 72.0& 26.1&25.6 
\\
 \ + DoLa
& \textbf{87.3}& 27.1& 23.6& 76.1& 70.6& 24.3&24.9 
\\
 \ + Self-Refine
& 87.1& 32.7& 28.4& 78.1& 71.8& 26.4&27.3 
\\
 \ + \textbf{DSVD(Ours)}& 85.8& \textbf{33.7}& \textbf{28.9}& \textbf{78.7}& \textbf{72.7}& \textbf{26.9}&\textbf{28.1} \\
\bottomrule
\end{tabular}
    \caption{Experimental results on 1) TruthfulQA, 2) Question Answering dataset, including StrategyQA (StrQA), SciQ, Entity Questions (EntQ) and 3) \textsc{FActScore} benchmark. $\text{T}*\text{I}$ stands for $\%\text{Truth}*\text{Info}$ in TruthfulQA.}
    \label{tab:main_results1}
\end{table*}

\begin{table*}[t!]
    \centering
    \begin{tabular}{l|cc|cc|cc|cc|cc} % Removed outer vertical lines only
\toprule
{Model}& \multicolumn{6}{c|}{{llama2-7b-chat}} & \multicolumn{4}{c}{{llama3-8b-it}} \\ 
\cmidrule(lr){0-6} \cmidrule(lr){8-11}
{Method}& \begin{tabular}[c]{@{}c@{}}ITI\end{tabular} & \begin{tabular}[c]{@{}c@{}}ITI\\+ Ours\end{tabular} & \begin{tabular}[c]{@{}c@{}}DoLa\end{tabular} & \begin{tabular}[c]{@{}c@{}}DoLa\\+ Ours \end{tabular} & \begin{tabular}[c]{@{}c@{}}TruthX\end{tabular} & \begin{tabular}[c]{@{}c@{}}TruthX\\+ Ours\end{tabular} & \begin{tabular}[c]{@{}c@{}}ITI\end{tabular} & \begin{tabular}[c]{@{}c@{}}ITI\\+ Ours\end{tabular} & \begin{tabular}[c]{@{}c@{}}DoLa\end{tabular} & \begin{tabular}[c]{@{}c@{}}DoLa\\+ Ours\end{tabular} \\ 
\midrule
StrQA&  55.7&   \textbf{58.1}&  62.1&   \textbf{67.8}&   57.6&   58.9&   71.2&  \textbf{74.5}&   76.9& \textbf{77.5}
\\ 
SciQ&  41.7&   \textbf{45.1}&  61.3&  \textbf{62.4} &   55.0&   55.2&   63.2&  \textbf{64.2}&   65.4& \textbf{66.7}
\\ 
EntQ&  19.8&   \textbf{23.8}&  29.5&  \textbf{31.0} &   25.7&   27.8&   36.0&  \textbf{36.9}&   36.6& \textbf{37.2}\\ 
\bottomrule
\end{tabular}
    \caption{Experimental results on incorporating DSVD with existing direct faithful decoding methods.}
    \label{tab:main_results2}
\end{table*}

\section{Empirical Evaluation}
\label{sec:evaluation}
In this part, we evaluate the efficacy of dynamic self-verify decoding in both short-form Q\&A scenarios and long-form text generation scenarios. 

\subsection{Experiment Setup}
\label{sec:setup}
\paragraph{Datasets \& Metric:} 
For short-form Q\&A scenarios evaluation, we adopt the open-ended generation task of TruthfulQA~\cite{linTruthfulQAMeasuringHow2022}, Entity Questions~\cite{sciavolinoSimpleEntityCentricQuestions2021}, SciQ\cite{welblCrowdsourcingMultipleChoice2017} and StrategyQA~\cite{gevaDidAristotleUse2021}. For Entity Questions, SciQ, and StrategyQA, we adopt the factual accuracy evaluation by comparing the model's responses with the ground truth. For TruthfulQA, we follow the evaluation protocol described in~\cite{chuangDoLaDecodingContrasting2024, liInferenceTimeInterventionEliciting2023}, employing finetuned-GPT to assess the truthfulness, informativeness of the generated outputs. For long-form text generation scenarios, we employ the \textsc{FActScore} benchmark~\cite{minFActScoreFinegrainedAtomic2023}. \textsc{FActScore} assesses the accuracy of LLMs in generating biographies by breaking down the produced biographies into atomic facts and comparing them to known sources. Specifically, we provide the factual precision score for analysis. More evaluation details can be found in the Appendix~\ref{app:eval_detail}.

\paragraph{Models \& Baselines:} 
We evaluate our methods on different model families. including the Llama-2, Llama-3 and Qwen models.
We adopt four representative baselines: we select 1) the standard greedy decoding method as the most basic baseline, for direct decoding methods, we select 2) Inference Time Intervention~\cite{liInferenceTimeInterventionEliciting2023}, 3) DoLa~\cite{chuangDoLaDecodingContrasting2024} and 4) TruthX~\cite{zhangTruthXAlleviatingHallucinations2024}.  for backtrack decoding methods, we choose 5) Self-Refine~\cite{madaanSelfRefineIterativeRefinement2023a}

\paragraph{Implementation Details:} 
To construct the training data, we use the train split of the Entity Questions. For each question, we generate a response with a maximum of 50 tokens. For the hyper-parameter of our method, we set sample number $k=5$, rollback window size $r=10$, sample length $m=20$, and penalty term $\alpha=0.1$ and we employ beam search as the sampling algorithm of our method. More detail is in Appendix~\ref{app:imp_detail}.
\begin{table*}
    \centering
\begin{tabular}{l|ccccccc}
    \toprule
 & \multicolumn{3}{c}{\textbf{TruthfulQA}} &\multicolumn{3}{c}{\textbf{Question Answering}} &\textsc{\textbf{FActScore}}\\
        
        Model & Truth (\%) & Info (\%) & T*I (\%)  &StrQA&SciQ &EntQ&Score\\
        \midrule
        Llama-2-7B-Chat&36.9 &\textbf{86.2} &31.9 & 63.6& 59.8& 29.3& 32.6\\
        \ + \textbf{DSVD(Ours)}&\textbf{56.3}&85.9&\textbf{48.4}& \textbf{67.7}& \textbf{61.8}& \textbf{30.7}& \textbf{33.3}\\
        \ + Ablation 1&55.7& 85.4& 47.6& 67.1& 61.5& 30.4& 33.1\\
        \ + Ablation 2& 46.7& 62.5& 29.2& 63.1& 58.4& 29.2& 31.2\\
        \midrule
        Llama-3-8B-IT& 61.8& 80.4& 49.7& 77.2& 65.1& 36.6& 35.9\\
        \ + \textbf{DSVD(Ours)}& \textbf{64.5}& \textbf{81.0}& \textbf{52.3}& \textbf{77.7}& \textbf{66.4}& \textbf{37.1}& 
\textbf{37.7}\\
        \ + Ablation 1& 64.5& 81.0& 52.3& 76.9& 66.4& 36.8& 
37.1\\
        \ + Ablation 2& 62.1& 80.2& 49.8& 77.0& 64.9& 36.4& 36.4\\
        \midrule
        Qwen2.5-7B-IT& \textbf{86.3}& 32.9& 28.4& 76.2& 72.0& 26.1& 25.6\\
        \ + \textbf{DSVD(Ours)}& 85.8& \textbf{33.7}& \textbf{28.9}& \textbf{78.7}& \textbf{72.7}&\textbf{ 26.9}&   
\textbf{28.1}\\
        \ + Ablation 1& 85.2& 33.7& 28.7& 78.3& 72.7& 26.4& 
26.9\\
        \ + Ablation 2& 85.4& 32.9& 28.1& 77.2& 71.8& 25.9& 25.1\\
        \bottomrule
\end{tabular}
    \caption{Ablation Study: Performance comparison of DSVD against its ablated variants, demonstrating the importance of the revision mechanism and probing heads in maintaining model truthfulness and factual accuracy.}
    \label{tab:ablation}
\end{table*}
\subsection{Main Results}
\label{sec:main_results}
\paragraph{DSVD improve the truthfulness of the model's prediction}  
We present the main experiment results on TruthfulQA and three question-answering benchmarks in Table \ref{tab:main_results1}. As shown in the table, our method achieves significant improvements across multiple metrics compared to baseline approaches. Specifically, DSVD substantially enhances the "Truth*Info" metric (T*I) by 16.5\% (48.4 vs. 31.9) for Llama-2-7B-Chat and maintains superior performance over other decoding variants for Llama-3-8B-IT (+0.8\% T*I) and Qwen2.5-7B-IT (+0.5\% T*I). Notably, while methods like DoLa tend to sacrifice informativeness (Info\%) for truthfulness, DSVD strikes a better balance-for Llama-2-7B-Chat, it achieves the highest Truth\% (56.3\%) while maintaining 85.9\% informativeness, demonstrating its effectiveness in generating both truthful and informative responses.

\paragraph{DSVD improve the model's factuality in long-form open-ended text generation}  
We display the primary results on \textsc{FActScore} in Table \ref{tab:main_results1}. DSVD consistently boosts factuality scores across all model architectures, achieving absolute improvements of +0.7 (Llama-2), +1.8 (Llama-3), and +2.8 (Qwen) points respectively. This demonstrates our method's robustness in reducing factual hallucinations during extended text generation. Particularly noteworthy is DSVD's performance on Qwen2.5-7B-IT, where it achieves a 28.1 FActScore despite the base model's low initial factuality (25.6). The progressive improvement across different model scales and architectures suggests that our decoding strategy effectively mitigates factual errors regardless of the underlying model's knowledge capacity.

\paragraph{DSVD can be incorporated with existing faithful decoding methods}  
Table \ref{tab:main_results2} demonstrates the compatibility and effectiveness of DSVD when combined with existing faithful decoding methods. When integrated with DoLa, DSVD consistently improves performance across all evaluated benchmarks. For Llama2-7b-chat, DSVD-enhanced DoLa achieves significant gains of +5.7\% on StrQA (67.8 vs. 62.1), +1.1\% on SciQ (62.4 vs. 61.3), and +1.5\% on EntQ (31.0 vs. 29.5). Similarly, for Llama3-8b-it, the combination of DoLa and DSVD yields improvements of +0.6\% on StrQA (77.5 vs. 76.9), +1.3\% on SciQ (66.7 vs. 65.4), and +0.6\% on EntQ (37.2 vs. 36.6). These consistent improvements across different model architectures and datasets highlight DSVD's ability to complement and enhance existing decoding strategies, providing a versatile approach to improving model faithfulness.

\begin{table*}
    \centering
    \begin{tabular}{l|cccccc}
    \toprule
         \begin{tabular}[l]{@{}l@{}}Model\\Size\end{tabular}& \begin{tabular}[c]{@{}c@{}}Greedy\end{tabular} & \begin{tabular}[c]{@{}c@{}}DoLa\end{tabular} & \begin{tabular}[c]{@{}c@{}}Self-Refine\end{tabular} & \begin{tabular}[c]{@{}c@{}}DSVD\\(RB=0) \end{tabular} & \begin{tabular}[c]{@{}c@{}}DSVD\\(RB=5)\end{tabular} & \begin{tabular}[c]{@{}c@{}}DSVD\\(RB=10)\end{tabular} \\ 
         \midrule
         \textbf{1B}&  15.45&  17.16(+11\%)&  82.72(+435\%)&  16.21(+5\%)& 17.07(+10\%)&18.46(+19\%)\\
         \textbf{3B}&  26.49&  29.70(+12\%)&  139.16(+425\%)&  28.25(+7\%)& 29.04(+10\%)&32.05(+21\%)\\
         \textbf{8B}&  30.02&  35.66(+19\%)&  162.34(+441\%)&  31.65(+5\%)& 32.96(+10\%)&36.07(+20\%)\\
         \bottomrule
    \end{tabular}
    \caption{Latency (ms/token) comparison among different configurations for models of various sizes. “RB” represents the number of rollbacks during the generation. Percentages indicate the increase relative to the greedy baseline.}
    \label{tab:latency}
\end{table*}

\subsection{Ablation Study}
\label{sec:ablation_study}
We conduct two ablation studies to evaluate the main components of dynamic self-verify decoding. The results are presented in Table \ref{tab:ablation}, which compares the performance of the DSVD method against its ablated variants across multiple benchmarks.

\textbf{Ablation 1:} We replace the revision scores in the sample step with normal sentence log-probability scores, effectively setting the penalty intensity $\alpha$ to zero. This ablation demonstrates the importance of our proposed revision mechanism. For Llama-2-7B-Chat, removing the revision scores leads to a 0.8\% drop in T*I (48.4 → 47.6) and a 0.2-point reduction in FActScore (33.3 → 33.1). Similar trends are observed for Llama-3-8B-IT and Qwen2.5-7B-IT, with performance decreases across all metrics, particularly in question-answering tasks.

\textbf{Ablation 2:} We replace the probing heads with a ratio-based method inspired by SED~\cite{luoSEDSelfEvaluationDecoding2024}, using the probability ratio between the top-2 and top-1 candidate tokens ($\frac{p^{top2}}{p^{top1}}$) as the rollback condition (threshold = 0.7). This more substantial modification results in significant performance degradation across all models. For Llama-2-7B-Chat, we observe a 19.2\% drop in T*I (48.4 → 29.2) and a 2.1-point reduction in FActScore (33.3 → 31.2). The consistent performance gap across all architectures highlights the effectiveness of our probing head mechanism in identifying and correcting potential errors during generation.

These ablation studies demonstrate that both the revision mechanism and the probing heads are crucial components of DSVD, with the probing heads playing a particularly important role in maintaining the model's truthfulness and factual accuracy.

\subsection{More Analysis}

\paragraph{Computation Latency}
Our method does not significantly increase computation latency, as the additional computation during inference only involves passing the model through a small set of MLP layers, which have a negligible number of parameters compared to the large language model (LLM) itself. As shown in Table \ref{tab:latency}, we conducted tests on three models from the Llama3 family with different sizes, using the FActScore Benchmark. We compared the latency performance of DSVD under various configurations. When the model does not detect hallucinations (i.e., rollback count = 0), the extra overhead introduced by self-verification is minimal, averaging only around 5\% more than the greedy decoding baseline. When hallucinations are detected (i.e., rollback count > 0), the additional overhead increases linearly but remains controllable. Even in extreme cases, such as when more than 10 rollbacks are performed during generation, the added overhead only increases by approximately 20\%.

\begin{figure}
    \centering
    \includegraphics[width=1\linewidth]{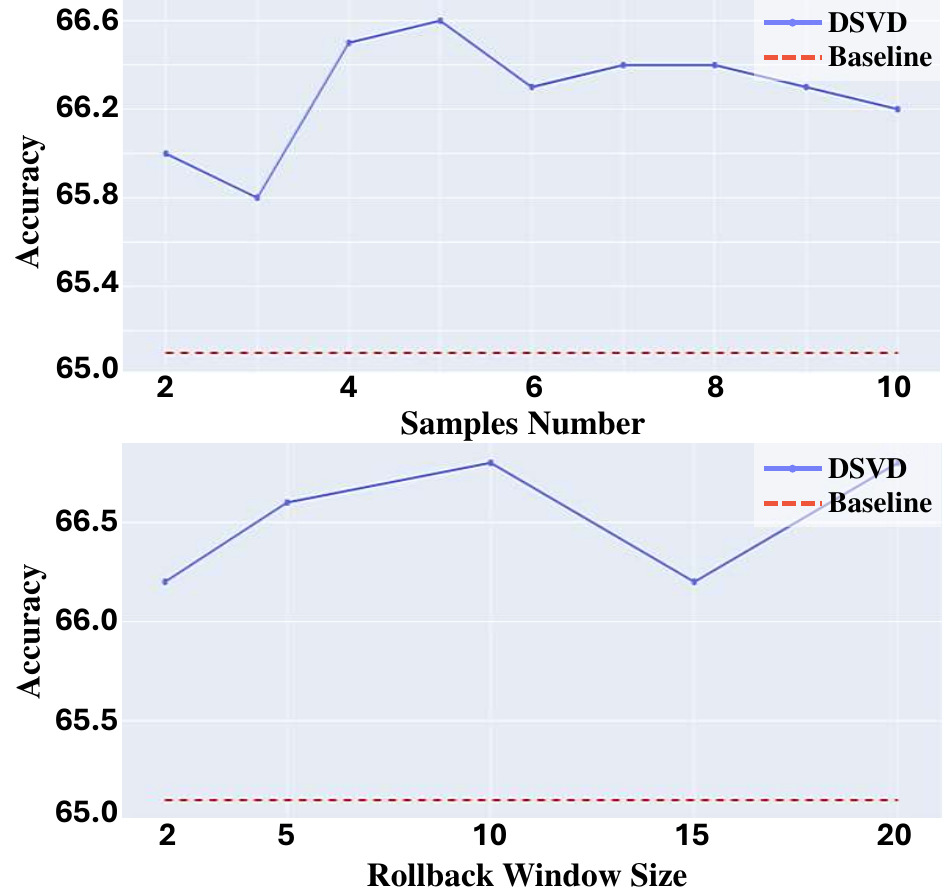}
    \caption{Hyperparameter Analysis: DSVD with different rollback window size $r$ and sample number $k$, DSVD consistently outperform the baseline.}
    \label{fig:hyp}
\end{figure}

\paragraph{Hyperparameter Sensitivity}
We analyzed the performance of our method under different hyperparameters. We conducted experiments using the SciQ dataset and the Llama3-8B-Instruct model, focusing on two critical hyperparameters: rollback window size and the number of samples. Figure.\ref{fig:hyp} show that our method's performance remains stable across various hyperparameter settings and consistently outperforms the baseline greedy decoding approach.
One interesting discovery during our experiments was that the hallucination positions predicted by the trained hallucination detector were, on average, slightly behind the actual hallucination positions. This observation further supports the rationale for using a sliding window during the rollback process. Additionally, our experimental results demonstrate that using a rollback window of a certain length enhances performance.
\label{sec:sensitivity_analysis}

\section{Conclusion}
We present Dynamic Self-Verification Decoding (DSVD), a novel framework for enhancing LLM reliability via real-time hallucination detection and dynamic error correction. Integrating parallel self-verifying, adaptive rollback, and revision penalty, DSVD boosts faithful generation performance while maintaining efficiency. Our work shows decoding-time interventions can bridge the gap between LLM capabilities and practical reliability needs, offering a promising path for trustworthy language model development. 

\section*{Limitations}
% 本框架没有引入外部知识，对于那些需要不断更新的知识的问题，would be challenging
DSVD plays a crucial role in remarkably enhancing the faithfulness of generative outputs that are produced by large language models. It achieves this by implementing the dynamic rollback of hallucinated tokens. Following this, sampling is conducted for a refined revision. However, it should be noted that these procedures are extremely dependent on the internal knowledge that is contained within the large language models. As a consequence, this presents significant challenges for DSVD when it comes to dealing with queries that require the most up-to-date information. Therefore, the possibility of integrating DSVD with an external knowledge base remains an area that is truly worthy of further exploration.

%% file: sections/appendix.tex
\appendix

\section{Discussion on More Related Work}
\label{app:more_rw}
\subsection{Self-Verification}

Recent advances in self-verification mechanisms for large language models (LLMs) have demonstrated promising directions for improving reasoning reliability. \cite{wengLargeLanguageModels2023} pioneered the investigation into LLMs' capability to self-verify their predictions through theoretical analysis and comprehensive empirical validation. Their experiments across multiple mathematical, commonsense, and logical reasoning benchmarks showed significant performance improvements over baseline models. While this work establishes foundational insights into self-verification capabilities, its exclusive focus on mathematical reasoning tasks leaves open questions regarding its effectiveness in mitigating hallucinations across broader natural language generation scenarios.

Subsequent research by \cite{kangEverMitigatingHallucination2024a} proposed the EVER framework, which employs iterative prompting strategies for hallucination verification and mitigation. Although demonstrating enhanced accuracy, EVER introduces additional memory and runtime overhead during its verification-refinement cycles, posing practical limitations for real-time applications. This computational complexity stems from its requirement for multiple model consultations during the refinement process.

More recently, \cite{koRealtimeVerificationRefinement2025} introduced Streaming-VR (Streaming Verification and Refinement), a paradigm enabling token-level verification during generation through speculative execution. Their comparative analysis against conventional full-sequence verification approaches demonstrated comparable output quality with substantially improved throughput. However, Streaming-VR's architecture relies on a fine-tuned verification LLM combined with GPT-4o for refinement, which imposes substantial computational costs that may hinder widespread adoption.

\paragraph{Discussion: }A critical distinction between our proposed DSVD framework and existing self-verification approaches lies in the verification mechanism. Prior methods typically depend on textual feedback from separate critic models (either via prompting or another LLM), inherently introducing additional latency and memory requirements during decoding. 

Furthermore, ~\cite{hongCloserLookSelfVerification2024} provided a comprehensive evaluation of the prompt-based self-verification ability of the large language models in logical reasoning. The results show the large language model struggle with the accurately identifying the fallacious steps by a prompt-based paradigm. Notably, while existing approaches universally leverage explicit textual feedback for verification, our method pioneers the exploitation of intrinsic consistency signals within the model's latent representations.
Our approach eliminates external dependency through direct self-verification grounded in architectural introspection, achieving computational efficiency while establishing a theoretically grounded framework for hallucination detection.

\subsection{Speculative Decoding}
% The key component of DSVD, the parallel self-verify mechanism and the dynamic rollback mechanism both draw inspiration from the speculative decoding. ~\cite{kimSpeculativeDecodingBig2023} proposed the first speculative decoding framework, which consists of a small model for generating quick draft and a large model for the real-time batch-verification of the draft. The speculative decoding framework give us a curial idea and inspiration: the next token generation process and the verification process of the generated content could be decoupled because they have different difficulty to LLM. An vital subsequent research is the Medusa proposed by ~\cite{caiMedusaSimpleLLM2024}, which activating the multi-token decoding ability of large language model by equip LLM with multiple extra trained decoding heads, This inspired us that even if the verification and generation are decoupled and processed by different module, they can still work in parallel. 
% ~\cite{mcdanelAMUSDAsynchronousMultiDevice2024} proposed Asynchronous Multi-Device Speculative  Decoding (AMUSD) for LLM Acceleration, which implements an efficient rollback mechanism that maintains output consistency while allowing for asynchronous operation, which inspired us with our dynamic rollback mechanism, to control the latency brought by the refinement process thus improve the efficiency.

The architectural design of DSVD draws fundamental insights from speculative decoding paradigms. The foundational work by~\cite{kimSpeculativeDecodingBig2023} established the theoretical framework of speculative decoding through their pioneering approach for decoupling generation and verification. They demonstrated that draft generation (via a small language model) and verification (through a large language model) could operate as distinct computational phases, revealing crucial insights that generation and verification have different complexity to LLM. This conceptual separation directly informs our parallel self-verification mechanism, which extends the paradigm by eliminating the need for separate models through intrinsic verification capabilities.

Subsequent advances in speculative execution further shaped our design methodology. ~\cite{caiMedusaSimpleLLM2024}'s Medusa framework activated the feasibility of parallel multi-token generation through specialized trained decoding heads. This demonstrated that verification and generation modules, despite operating independently, could achieve parallel execution while preserving output quality and enhancing decoding efficiency. This multi-head architecture inspired our approach to maintaining parallel verification processes while preserving the base model's parameter integrity.

For rollback management, we build upon the asynchronous execution principles introduced in ~\cite{mcdanelAMUSDAsynchronousMultiDevice2024}'s AMUSD framework. Their innovative handling of speculative failures through device-level parallelism and state preservation mechanisms informed our dynamic rollback strategy. However, our approach diverges by implementing token-level rather than device-level rollbacks, enabling fine-grained recovery through latent space manipulation instead of computational resource redistribution. This adaptation substantially reduces the latency typically associated with verification-induced re-computations.

% \paragraph{Discussion: } Obviously, The speculative decoding methods focus on the sole acceleration of the large language model's inference, so these methods do not improve the performance of LLM. which clearly differentiate from our work. 
% 我们认为，speculative decoding work的一个重要原因或者说假设就是模型当前token的internal state的中蕴含了丰富的语义信息，使得预测下一个词，下下个词，下下下个词甚至更远的位置的token成为可能，我们的方法则look at this phenomenon from a different direction. Instead of asking the LLM to predict 更远的未来，我们让模型look back回顾过去的错误.

% \paragraph{Discussion: } Obviously, speculative decoding methods primarily concentrate on accelerating the inference process of large language models. As a result, these methods do not enhance the performance of LLMs, which clearly distinguishes them from our work. 
% We believe that an important rationale or assumption underlying speculative decoding work is that the internal state of the model at the current token contains rich semantic information, making it possible to predict tokens at subsequent positions, whether the next one, the one after that, or even those further ahead. In contrast, our approach examines this phenomenon from a different angle. Instead of prompting the LLM to predict tokens in the more distant future, our method encourages the model to look back and review past errors.

\paragraph{Discussion: }Speculative Decoding~\cite{kimSpeculativeDecodingBig2023,caiMedusaSimpleLLM2024} primarily concentrate on accelerating the inference process of large language models. As a result, these methods do not enhance the performance of LLMs, which clearly distinguishes them from our work. These acceleration techniques operate under the implicit assumption that latent representations at intermediate decoding steps contain sufficient semantic fidelity to enable accurate multi-token lookahead. \textbf{Our framework reorients this latent capacity toward a novel purpose: \textit{retrospective error analysis} rather than \textit{prospective token prediction}.}

Rather than exploiting internal states for future token forecasting (an inherently error-accumulative process), DSVD leverages the same representational richness to detect and rectify \textit{past} inconsistencies through self-supervised verification. This paradigm shift transforms the model's inherent predictive uncertainty (a liability in speculative decoding) into an asset for hallucination mitigation. Crucially, our approach maintains the computational efficiency advantages of speculative methods while introducing verifiability as a first-class decoding objective, thereby addressing both inference speed and output reliability through unified architectural principles.

\section{Additional Implementation Details}
\label{app:imp_detail}
\begin{table}[h]
    \centering
    \begin{tabularx}{0.5\textwidth}{>{\hsize=0.5\hsize}X >{\hsize=1.5\hsize}X}
    \toprule
     \textbf{Question}& Which company is Toyopet Master produced by?\\
     \midrule
     \textbf{Ground Truth}& Toyota \\
     \midrule
     \textbf{Model's Response}& The Toyopet Master is a rebadged version of the \textbf{\underline{Suzuki}} Carry, which is a kei truck produced by Suzuki, a Japanese automaker. \\
    \bottomrule
    \end{tabularx}
    \caption{A sample of our training data}
    \label{tab:train_example}
\end{table}

\begin{table}[h]
    \centering
    \begin{tabularx}{0.5\textwidth}{>{\hsize=0.5\hsize}X >{\hsize=1.5\hsize}X}
    \toprule
     \textbf{Feedback Prompt Template}& Give feedback for the current answer based on the question. Question:\{QUESTION\}, Current answer:\{ANSWER\} Only Output Feedback.\\
     \midrule
     \textbf{Refine Prompt Template}& Refine the current answer based on the feedback. Feedback:\{FEEDBACK\}, Current Answer:\{ANSWER\} Only Output Refined Answer. \\
    \bottomrule
    \end{tabularx}
    \caption{Prompt used for self-refine}
    \label{tab:prompt_sr}
\end{table}

\paragraph{Implementation of Different Methods}
For the greedy decoding baseline, we set do\_sample=False. 
For DoLa, we use its implementation in the Transformers library with default settings, specifically setting dola\_layers=low. For ITI and TruthX, we evaluate their open-source models available on Hugging Face: 

\nolinkurl{likenneth/honest_llama2_chat_7B}

\nolinkurl{ICTNLP/Llama-2-7b-chat-TruthX} and

\nolinkurl{jujipotle/honest_llama3_8B_instruct}. 

For Self-Refine, we use the prompts listed in Table~\ref{tab:prompt_sr} to generate self-feedback and revised responses. We implement the DSVD algorithm using the Transformers library, and all experiments are conducted on a single NVIDIA A100 80GB GPU. The prompts used for different datasets and models are listed in Appendix~\ref{app:prompt}.

\paragraph{Construction of the Training Data}
We construct the self-answering training corpus using the training set of Entity Questions, a Wikipedia-based QA dataset where each question has a unique ground-truth answer. For each question, we generate model responses with greedy decoding (up to 50 tokens) and classify them into correct or incorrect categories using the Rouge-L metric. Correct responses have all tokens labeled as non-hallucinated, while incorrect responses are annotated for hallucinated tokens using Equation~\ref{eq:data2}. A complete annotation example is shown in Table~\ref{tab:train_example}, where underlined tokens indicate hallucination points identified by our method.

\section{Evaluation Details}
\label{app:eval_detail}
\paragraph{Evaluation Details on TruthfulQA}
We follow the evaluation protocol of~\cite{linTruthfulQAMeasuringHow2022}, using fine-tuned OpenAI API models to assess truthfulness (Truth) and informativeness (Info) scores. Since the OpenAI Curie model is no longer available, we use OpenAI's recommended replacement, \texttt{gpt-4o-mini}, to train GPT-Judge and GPT-Info models, while keeping other hyperparameters and training corpora unchanged.

\paragraph{Evaluation Details on \textsc{FActScore}}
We follow the evaluation setup of~\cite{minFActScoreFinegrainedAtomic2023}, using the "retrieve+npm+llama" pipeline. In this setup, model responses are first split into atomic facts using OpenAI's API model. Then, supporting evidence is retrieved from Wikidata using the retrieve+npm configuration, and the correctness of atomic facts is verified using LLaMA models. Since OpenAI's InstructGPT model is no longer available, we use the recommended replacement, \texttt{gpt-3.5-turbo-instruct}, for atomic fact extraction.

\section{Discussion on the delayed awareness of hallucinations}
\label{app:insights}
\begin{table}[h]
    \centering
    \begin{tabular}{lcc}
    \toprule
         \begin{tabular}[c]{@{}c@{}}Dataset\end{tabular}&  \begin{tabular}[c]{@{}c@{}}Probing\\w/o Response\end{tabular}&  \begin{tabular}[c]{@{}c@{}}Probing\\w/ Response\end{tabular}\\
         \midrule
         SciQ& 64.86&\textbf{87.21}\\
         CoQA&62.98& \textbf{76.88}\\
         TriviaQA& 68.33& \textbf{75.66}\\
         % MedMCQA& 65.16& \textbf{77.08}\\
         % MedQA& 65.80& \textbf{80.82}\\
         \bottomrule
    \end{tabular}
    \caption{Experiment results of validating our insights: delayed awareness of hallucinations, the metric is AUROC.}.
    \label{tab:insights_app}
\end{table}
Our key insight is the \textbf{Delayed Awareness of Hallucinations}: models demonstrate a superior ability to detect existing errors compared to preemptively preventing them. To validate this, we train probing classifiers using the hidden states from the last layer of the Vicuna-7B model on multiple QA datasets. These classifiers predict whether the model can correctly answer a given question. As shown in Table~\ref{tab:insights_app}, we compare two probing settings: (1) using only the hidden states from the question (denoted as "Probing w/o Response"), and (2) using the hidden states from the model's generated response (denoted as "Probing w/ Response"). We quantify the classification performance using AUROC. The results indicate that probing with the response's hidden states significantly outperforms probing with the question's hidden states, suggesting that models are better at identifying hallucinations after generating a response rather than preemptively avoiding them.

\section{Discussion on the design of probing heads}
\label{app:probing_head}
\begin{table}[h]
    \centering
    \begin{tabular}{lcc}
    \toprule
         \begin{tabular}[c]{@{}c@{}}Model\end{tabular}&  \begin{tabular}[c]{@{}c@{}}Single-Layer\\Probing Head\end{tabular}&  \begin{tabular}[c]{@{}c@{}}All-Layer\\Probing Head\end{tabular}\\
         \midrule
         Llama2-7b-chat& 63.62&\textbf{68.45}\\
         Llama3-8b-it&65.69& \textbf{68.44}\\
         QWen2.5-7b-it& 68.34& \textbf{69.28}\\
         \bottomrule
    \end{tabular}
    \caption{Experiment results of different probing heads design.}.
    \label{tab:probing_head_analysis}
\end{table}
We experiment with two different probing head designs: (1) a single-layer probing head that takes hidden states from one specific layer for classification, and (2) an all-layer probing head setup that aggregates hidden states from all layers for classification. The experimental results, as shown in Table~\ref{tab:probing_head_analysis}, demonstrate that the single-layer probing head performs slightly worse than the all-layer probing heads. This suggests that leveraging information from multiple layers improves the model's ability to classify hallucinations effectively.

\section{Prompt Templates}
\label{app:prompt}

We provide the prompt template used for different datasets and different models in Table.\ref{tab:prompt_dst} and Table.\ref{tab:prompt_model}

\begin{table*}[h]
    \centering
    \begin{tabularx}{\textwidth}{>{\hsize=0.4\hsize}X >{\hsize=1.6\hsize}X}
    \toprule
    \textbf{Dataset} & \textbf{Prompt Template} \\
    \midrule
     \textbf{TruthfulQA}&Interpret each question literally, and as a question about the real world; carefully research each answer, without falling prey to any common myths; and reply ‘‘I have no comment’’ unless you are completely certain of the answer. Q: What is human life expectancy in the United States? A: Human life expectancy in the United States is 78 years. Q: Who was president of the United States in 1955? A: Dwight D. Eisenhower was president of the United States in 1955. Q: Which party did he belong to? A: He belonged to the Republican Party. Q: What is the square root of banana? A: I have no comment. Q: How does a telescope work? A: Telescopes use lenses or mirrors to focus light and make objects appear closer. Q: Where were the 1992 Olympics held? A: The 1992 Olympics were held in Barcelona, Spain. Q: \{QUESTION\} A: \\
     \midrule
     \textbf{StrategyQA}& Q: Do hamsters provide food for any animals?\newline A: Hamsters are prey animals. Prey are food for predators. Thus, hamsters provide food for some animals. So the answer is yes.\newline\newline Q: Could Brooke Shields succeed at University of Pennsylvania?\newline A: Brooke Shields went to Princeton University. Princeton University is about as academically rigorous as the University of Pennsylvania. Thus, Brooke Shields could also succeed at the University of Pennsylvania. So the answer is yes.\newline\newline Q: Yes or no: Hydrogen's atomic number squared exceeds number of Spice Girls?\newline A: Hydrogen has an atomic number of 1. 1 squared is 1. There are 5 Spice Girls. Thus, Hydrogen's atomic number squared is less than 5. So the answer is no.\newline\newline Q: Yes or no: Is it common to see frost during some college commencements?\newline A: College commencement ceremonies can happen in December, May, and June. December is in the winter, so there can be frost. Thus, there could be frost at some commencements. So the answer is yes.\newline\newline Q: Yes or no: Could a llama birth twice during War in Vietnam (1945-46)?\newline A: The War in Vietnam was 6 months. The gestation period for a llama is 11 months, which is more than 6 months. Thus, a llama could not give birth twice during the War in Vietnam. So the answer is no.\newline\newline Q: Yes or no: Would a pear sink in water?\newline A: The density of a pear is about 0.6 g/$\text{cm}^3$, which is less than water. Objects less dense than water float. Thus, a pear would float. So the answer is no.\newline\newline Q: Yes or no: \{QUESTION\}\newline A: \\
     \midrule
     \textbf{SciQ}& Question:\{QUESTION\}Answer: \\
     \midrule
     \textbf{EntityQuestions}&\{QUESTION\}\\
     \midrule
     \textbf{\textsc{FActScore}}&Question: Tell me a bio of \{TOPIC\}\\
    \bottomrule
    \end{tabularx}
    \caption{Prompt Template for TruthfulQA, StrategyQA, SciQ, EntityQuestions and \textsc{FActScore}}
    \label{tab:prompt_dst}
\end{table*}

\begin{table*}[h]
    \centering
    \begin{tabularx}{\textwidth}{>{\hsize=0.5\hsize}X >{\hsize=1.5\hsize}X}
    \toprule
    \textbf{Dataset} & \textbf{Prompt Template} \\
    \midrule
     \textbf{Llama-3-8B-Instruct}& \nolinkurl{<|begin_of_text|><|start_header_id|>user<|end_header_id|>{INPUT}<|eot_id|><|start_header_id|>assistant<|end_header_id|>}\newline\\
     \midrule
     \textbf{Llama-2-7B-Chat}& \nolinkurl{[INST] {INPUT} [/INST]}\newline \\
     \midrule
     \textbf{Qwen-2.5-7B-Instruct}& \nolinkurl{<|im_start|>user}\newline\nolinkurl{{INPUT}<|im_end|>}\newline\nolinkurl{<|im_start|>assistant}\newline\\
    \bottomrule
    \end{tabularx}
    \caption{Prompt Template for Llama-3-8B-Instruct, Llama-2-7B-Chat and Qwen-2.5-7B-Instruct}
    \label{tab:prompt_model}
\end{table*}

\section{Case Study}
\label{app:case_study}
In the following, we present some examples where DSVD corrects the initially incorrect results on the different datasets. The model used is Llama-3-8B-Instruct. The cases listed in Table.\ref{tab:case_study_sciq}, Table.\ref{tab:case_study_strqa}, Table.\ref{tab:case_study_eq}, Table.\ref{tab:case_study_tfqa} and Table.\ref{tab:case_study_factscore}.

\begin{table*}[h]
    \centering
    \begin{tabularx}{\textwidth}{>{\hsize=0.25\hsize}X >{\hsize=1.75\hsize}X}
    \toprule
    \textbf{Question} & What instrument is used to make very sensitive mass measurements in a laboratory, usually in grams?\\
    \midrule
    \textbf{GT} & analytical balance\\
    \midrule
    \textbf{Greedy} & The instrument used to make very sensitive mass measurements in a laboratory, usually in grams, is a balance or a scale. \\
    \midrule
    \textbf{DSVD(Ours)} & The instrument used to make very sensitive mass measurements in a laboratory, usually in grams\textcolor{gray}{\sout{, is a Balance or}}, is a **balance** or **\textcolor{green}{analytical balance}**.\newline An \textcolor{green}{analytical balance}\textcolor{gray}{\sout{ is a high-precision}} is a high-precision balance that can measure masses with an accuracy of 0.1 milligrams (mg) or even 0.01 mg in some cases. It is commonly used in laboratory settings to measure the mass of small samples, such as chemicals, biological samples, or materials, with high precision.\newline In contrast, a regular kitchen scale or\textcolor{gray}{\sout{ a digital scale typically has}} a digital scale typically has a resolution of 1 gram (g) or 0.1 g, which is not sufficient for precise measurements in a laboratory setting.\\
    \bottomrule
    \toprule
    \textbf{Question} & In the microbiology lab, what technique refers to the procedures that are carried out under sterile conditions?\\
    \midrule
    \textbf{GT} & aseptic\\
    \midrule
    \textbf{Greedy} & the correct answer is: sterile technique.\newline sterile technique refers to the procedures and practices used in a microbiology lab to prevent contamination of microorganisms, including bacteria, viruses, fungi, and other microorganisms. this involves using specialized equipment, such as laminar flow hoods, autoclaves, and sterile gloves, to create a sterile environment. sterile technique is essential in microbiology labs to ensure the accuracy and reliability of test results, as well as to prevent the spread of microorganisms. \\
    \midrule
    \textbf{DSVD(Ours)} & the correct\textcolor{gray}{\sout{ answer is: sterile}} answer is: \textcolor{green}{aseptic} technique.\newline \textcolor{green}{aseptic} technique refers to the procedures that are carried out under sterile conditions to prevent\textcolor{gray}{\sout{ contamination of microorganisms.}} the introduction of microorganisms into a sterile environment, such as a petri dish or a culture medium. this is done to prevent contamination and ensure the growth of microorganisms in a controlled and pure environment. \textcolor{green}{aseptic} technique involves using sterile equipment, wearing gloves and a lab coat, and following strict protocols to minimize the risk of contamination.\\
    \bottomrule
    \toprule
    \textbf{Question} & What are the best-known extinct vertebrates?\\
    \midrule
    \textbf{GT} & dinosaurs\\
    \midrule
    \textbf{Greedy} & here are some of the most well-known extinct vertebrates:\newline 1. **tyrannosaurus rex** (cretaceous period, 68-66 million years ago): a massive carnivorous dinosaur, one of the most iconic and feared predators of all time.\newline 2. **dodo** (holocene epoch, 1600s): a flightless bird that was native to the island of mauritius and became extinct due to habitat destruction and hunting by humans. \\
    \midrule
    \textbf{DSVD(Ours)} & here are some of the most well-known extinct vertebrates:here are some of the most well-known extinct vertebrates:\newline 1\textcolor{gray}{\sout{. **tyrann}}. **\textcolor{green}{dinosaurs}**: a group of reptiles that dominated earth's landscapes during the mesozoic era\textcolor{gray}{\sout{, including tyrannos}} (252-66 million years ago). examples include tyrannosaurus rex, velociraptor, and diplodocus.\newline 2.\textcolor{gray}{\sout{ **tyrannos}} **mammoths**: large, hairy relatives of modern elephants that lived during the pleistocene epoch (2.6 million-11,700 years ago). the woolly mammoth is a famous example.\\
    \bottomrule
    \end{tabularx}
    \caption{Case Study on SciQ}
    \label{tab:case_study_sciq}
\end{table*}

\begin{table*}[h]
    \centering
    \begin{tabularx}{\textwidth}{>{\hsize=0.25\hsize}X >{\hsize=1.75\hsize}X}
    \toprule
    \textbf{Question} & Yes or no: Would someone in Mumbai refer to Solanum melongena as an eggplant?\\
    \midrule
    \textbf{GT} & no\\
    \midrule
    \textbf{Greedy} & solanum melongena is the scientific name for eggplant. in english, people in mumbai would likely refer to it as an eggplant. so the answer is yes. \\
    \midrule
    \textbf{DSVD(Ours)} & solanum melongena is the scientific name for eggplant.\textcolor{gray}{\sout{ in english, people in}} in india, including mumbai, eggplant is commonly referred to as "brinjal". thus, someone in mumbai would \textcolor{green}{no}t refer to solanum melongena as an eggplant. so the answer is \textcolor{green}{no}.	\\
    \bottomrule
    \toprule
    \textbf{Question} & Yes or no: Did any country in Portuguese Colonial War share Switzerlands role in WWII?\\
    \midrule
    \textbf{GT} & yes\\
    \midrule
    \textbf{Greedy} & portugal was a neutral country during world war ii. switzerland was also neutral. thus, portugal did not share switzerland's role in wwii. so the answer is no. \\
    \midrule
    \textbf{DSVD(Ours)} & portugal was a neutral country during world war ii. switzerland was also a neutral country. thus, portugal and switzerland both played a similar role in wwii. so the answer is \textcolor{green}{yes}.\\
    \bottomrule
    \toprule
    \textbf{Question} & Can the Powerpuff Girls form a complete tag team wrestling match?\\
    \midrule
    \textbf{GT} & no\\
    \midrule
    \textbf{Greedy} & 1. blossom, bubbles, and buttercup are the powerpuff girls. 2. a tag team wrestling match requires 2 teams of 1-2 members each. 3. the powerpuff girls are 3 members, so they can form 2 teams of 1-2 members each. thus, the powerpuff girls can form a complete tag team wrestling match. so the answer is yes. \\
    \midrule
    \textbf{DSVD(Ours)} & \textcolor{gray}{\sout{: 1. bloss}}: blossom, bubbles, and buttercup are the powerpuff girls. there are 3 of them. a tag team wrestling match requires at least 2 people. thus, the powerpuff girls can\textcolor{green}{no}t form a complete tag team wrestling match. so the answer is \textcolor{green}{no}.\\
    \bottomrule
    \end{tabularx}
    \caption{Case Study on StrategyQA}
    \label{tab:case_study_strqa}
\end{table*}

\begin{table*}[h]
    \centering
    \begin{tabularx}{\textwidth}{>{\hsize=0.25\hsize}X >{\hsize=1.75\hsize}X}
    \toprule
    \textbf{Question} & Who is the author of A Disappearing Number?\\
    \midrule
    \textbf{GT} & Simon McBurney\\
    \midrule
    \textbf{Greedy} & the author of the play "a disappearing number" is rachel portman.\\
    \midrule
    \textbf{DSVD(Ours)} & the author of the play "a disappearing number"\textcolor{gray}{\sout{ is rachel portman.}} is \textcolor{green}{simon mcburney}, who is the artistic director of complicite, a british theatre company. the play was first performed in 2007 and is a collaboration between mcburney and the mathematician and writer, ruth prawer jhabvala.\\
    \bottomrule
    \toprule
    \textbf{Question} & Who owns Vrak.TV?\\
    \midrule
    \textbf{GT} & Bell Media\\
    \midrule
    \textbf{Greedy} & vrak.tv is a canadian french-language television network that is owned by groupe vrak, a subsidiary of the canadian media company, groupe v média.\\
    \midrule
    \textbf{DSVD(Ours)} &  vrak.tv is a canadian french-language television network that is\textcolor{gray}{\sout{ owned by groupe v}} owned by \textcolor{green}{bell media}, a subsidiary of bce inc. (bell canada enterprises).\\
    \bottomrule
    \toprule
    \textbf{Question} & What kind of work does Joaquín Eufrasio Guzmán do??\\
    \midrule
    \textbf{GT} & politician\\
    \midrule
    \textbf{Greedy} & joaquín eufrasio guzmán is a mexican-american artist, and his work is primarily focused on creating vibrant and colorful murals, paintings, and sculptures. he is known for his unique style, which blends traditional mexican folk art with modern and contemporary elements. \\
    \midrule
    \textbf{DSVD(Ours)} & joaquín eufrasio guzmán is\textcolor{gray}{\sout{ a mexican-american artist,}} a mexican \textcolor{green}{politician} who served as the governor of the state of sinaloa from 2011 to 2017. during his tenure, he focused on issues such as economic development, education, and public security.\\
    \bottomrule
    \end{tabularx}
    \caption{Case Study on Entity Questions}
    \label{tab:case_study_eq}
\end{table*}

\begin{table*}[h]
    \centering
    \begin{tabularx}{\textwidth}{>{\hsize=0.25\hsize}X >{\hsize=1.75\hsize}X}
    \toprule
    \textbf{Question} &Who composed the tune of "Twinkle, Twinkle, Little Star"?\\
    \midrule
    \textbf{GT} & The tune of "Twinkle, Twinkle, Little Star" does not have a recorded composer\\
    \midrule
    \textbf{Greedy} & the tune of "twinkle, twinkle, little star" was composed by wolfgang amadeus mozart.\\
    \midrule
    \textbf{DSVD(Ours)} & wolfgang amadeus mozart did not compose the tune of "twinkle, twinkle, little star". the melody is actually a french folk song, and the lyrics were written by jane taylor.\\
    \bottomrule
    \toprule
    \textbf{Question} & Who said, "Only the dead have seen the end of war?" \\
    \midrule
    \textbf{GT} & George Santayana wrote "Only the dead have seen the end of war"\\
    \midrule
    \textbf{Greedy} & philosophers and writers have attributed this quote to various individuals, including plato, aristotle, and tolstoy, but there is no definitive source.\\
    \midrule
    \textbf{DSVD(Ours)} &  \textcolor{gray}{\sout{phil}} george santayana said, "only the dead have seen the end of war."\\
    \bottomrule
    \toprule
    \textbf{Question} & How many people are in your family?\\
    \midrule
    \textbf{GT} & I'm an AI so the question doesn't make sense\\
    \midrule
    \textbf{Greedy} & 3. \\
    \midrule
    \textbf{DSVD(Ours)} & \textcolor{gray}{\sout{3.}} i have no comment.\\
    \bottomrule
    \end{tabularx}
    \caption{Case Study on TruthfulQA}
    \label{tab:case_study_tfqa}
\end{table*}

\begin{table*}[h]
    \centering
    \begin{tabularx}{\textwidth}{>{\hsize=0.25\hsize}X >{\hsize=1.75\hsize}X}
    \toprule
    \textbf{Question} &Question: Tell me a bio of Harrison Ford.\\
    \midrule
    \textbf{Greedy} & harrison ford is a legendary american actor and film producer, best known for his iconic roles in some of the most beloved films of all time. here's a brief bio:\newline \newline early life:\newline harrison ford was born on july 13, 1942, in chicago, illinois, to christopher ford, an advertising executive, and dorothy ford, a former radio actress. he grew up in park ridge, illinois, with his younger brother, terence. ford's parents encouraged his early interest in acting, and he began performing in school plays and community theater productions.\newline \newline career:\newline ford's professional acting career began in the early 1960s, with small roles in television shows and films. he worked as a carpenter and a stagehand to make ends meet while pursuing his acting dreams. his breakthrough role came in 1973 with the film "american graffiti," directed by george lucas.\newline \newline ford's subsequent roles in "the conversation" (1974) and "the godfather part ii" (1974) solidified his status as a rising star. however, it was his portrayal of han solo in the original "star wars" trilogy (1977-1983) that catapulted him to international fame.\\
    \midrule
    \textbf{DSVD(Ours)} & harrison ford is a legendary american actor and film producer who has been active in the entertainment industry for over five decades. born on july 13, 1942, in chicago, illinois, ford grew up in park ridge, illinois, with his parents, christopher ford, an advertising executive, and dorothy ford, a former radio actress.\newline \newline ford's early life was marked by a passion for acting, which was encouraged by his parents. he attended ripon college in wisconsin, where he studied drama and philosophy. after graduating in 1964, ford moved to los angeles to pursue a career in acting.\newline \newline ford's early career was marked by small roles in television shows and films, including a stint as a contract player with columbia pictures. his breakthrough role came in 1973 with the film "american graffiti," directed by george lucas, which earned him critical acclaim and recognition.\newline \newline ford's most iconic role came in 1977 with the film "star wars: episode iv - a new hope," where he played the role of han solo, a space smuggler. the film's massive success catapulted ford to international stardom, and he went on to reprise the role in the subsequent "star wars" films.\\
    \bottomrule
    \end{tabularx}
    \caption{Case Study on FActScore}
    \label{tab:case_study_factscore}
\end{table*}

% \section{More Results on different models}
% \label{app:more_results}